\documentclass[review]{elsarticle}
\usepackage{CJK}
\usepackage{indentfirst}
\usepackage{amsmath}
\usepackage{lineno}   
\usepackage[colorlinks,
 linkcolor=black,
 anchorcolor=black,
 citecolor=black
 ]{hyperref}
\journal{Journal of Food Engineering}

\begin{document}

\begin{frontmatter}

\title{Deep Learning-Based Food Calorie Estimation Method in Dietary Assessment\tnoteref{mytitlenote}}


\author[mymainaddress]{Yanchao Liang\corref{mycorrespondingauthor}}
\ead{yc\_liang@qq.com}

\author[mymainaddress]{Jianhua Li}
\cortext[mycorrespondingauthor]{Corresponding author}
\ead{jhli@ecust.edu.cn}

\address[mymainaddress]{School of Information Science and Engineering, East China University of Science and Technology, China}

\begin{abstract}
Obesity treatment requires obese patients to record all food intakes per day. Computer vision has been applied to estimate calories from food images. In order to increase detection accuracy and reduce the error of volume estimation in food calorie estimation, we present our calorie estimation method in this paper. To estimate calorie of food,  a top view and side view are needed. Faster R-CNN is used to detect each food and calibration object. GrabCut algorithm is used to get each food's contour. Then each food's volume is estimated by volume estimation formulas. Finally we estimate each food's calorie. And the experiment results show our estimation method is effective. 
\end{abstract}

\begin{keyword}
\texttt{computer vision}\sep \texttt{calorie estimation}\sep \texttt{deep learning} 
\end{keyword}

\end{frontmatter}

\section{Introduction}

Obesity is a medical condition in which excess body fat has accumulated to the extent that it may have a negative effect on health. People are generally considered obese when their Body Mass Index(BMI) is over 30 ${kg/m^2}$. High BMI is associated with the increased risk of diseases, such as heart disease, type two diabetes, etc\cite{BMI}. Unfortunately, more and more people will meet criteria for obesity. 
The main cause of obesity is the imbalance between the amount of food intake and energy consumed by the individuals. Conventional dietary assessment methods include food diary, 24-hour recall, and food
frequency questionnaire (FFQ)\cite{FFQ}, which requires obese patients to record all food intakes per day. In most cases, patients do have troubles in estimating the amount of food intake because they are unwillingness to record or lack of related nutritional information. While computer vision-based measurement methods were applied to estimate calories from food images which includes calibration objects; obese patients have benefited a lot from these methods.

In recent years, there are a lot of methods based on computer vision proposed to estimate calories\cite{circle_plate,collaboration_card,ebutton,mobile_cloud}. For these methods, the accuracy of estimation result is determined by two main factors: object detection algorithm and volume estimation method. In the aspect of object detection, classification algorithms like Support Vector Machine(SVM)\cite{SVM} are used to recognize food’s type in general conditions. In the aspect of volume estimation, the calibration of food and the volume calculation are two key issues. For example, when using a circle plate\cite{circle_plate} as a calibration object, it is detected by ellipse detection; and the volume of food is estimated by applying corresponding shape model. Another example is using people’s thumb as the calibration object, the thumb is detected by color space conversion\cite{RGB2YCBCR}, and the volume is estimated by simply treating the food as a column. However, thumb’s skin is not stable and it is not guaranteed that each person’s thumb can be detected. The involvement of human's assistance\cite{collaboration_card} can improve the accuracy of estimation but consumes more time, which is less useful for obesity treatment. After getting food’s volume, food's calorie is calculated by searching its density in food density table\cite{density_table} and energy in nutrition table\footnote{\url{http://www.hc-sc.gc.ca/fn-an/nutrition/fiche-nutri-data/nutrient_value-valeurs_nutritives-tc-tm-eng.php}}. Although these methods mentioned above have been used to estimate calories, the accuracy of detection and volume estimation still needs to be improved.

In this paper, we propose our calorie estimation method. This method takes two food images as its inputs: a top view and a side view; each image includes a calibration object which is used to estimate image’s scale factor. Food(s) and calibration object are detected by object detection method called Faster R-CNN and each food’s contour is obtained by applying GrabCut algorithm. After that, we estimate each food's volume and calorie. 

\section{Material and Methods}
\label{system}
\subsection{Calorie Estimation Method Based On Deep Learning}
\begin{figure}[htb!]
\centering
\includegraphics[width=\textwidth]{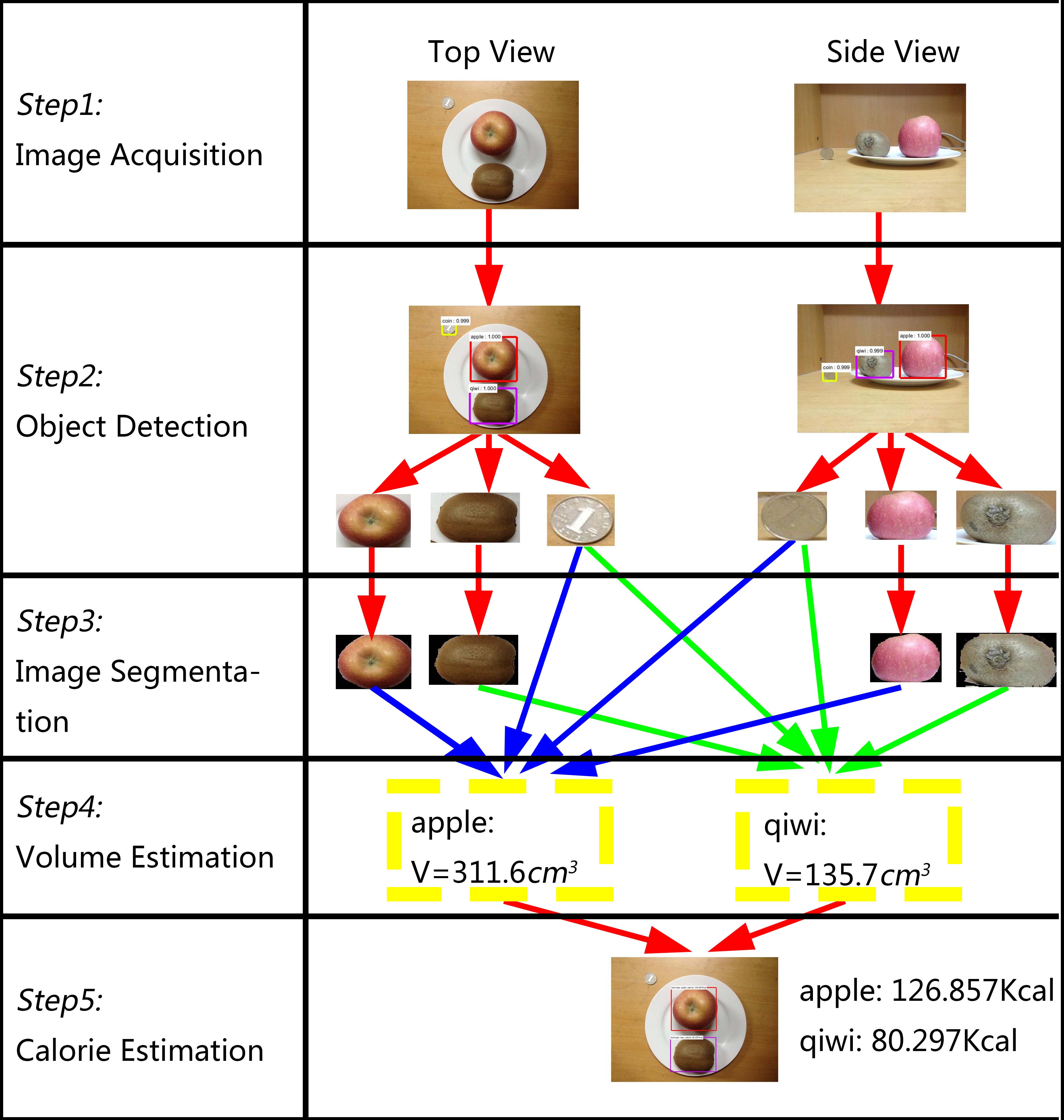}
\caption{Calorie Estimation Flowchart}
\label{calorie estimation}
\end{figure}

Figure \ref{calorie estimation} shows the flowchart of the proposed method. Our method includes 5 steps: image acquisition, object detection, image segmentation volume estimation and calorie estimation. To estimate calories, it requires the user to take a top view and a side view of the food before eating with his/her smart phone. Each images used to estimate must include a calibration object; in our experiments, we use One Yuan coin as a reference. 
In order to get better results, we choose to use Faster Region-based Convolutional Neural Networks (Faster R-CNN)\cite{fasterrcnn} to detect objects and GrabCut \cite{grabcut} as segmentation algorithms. 
\subsection{Deep Learning based Objection detection}
We do not use semantic segmentation method such as Fully Convolutional Networks (FCN)\cite{fcn} but choose to use Faster R-CNN. Faster R-CNN is a framework based on deep convolutional networks. It includes a Region Proposal Network (RPN) and an Object Detection Network\cite{fasterrcnn}. When we put an image with RGB channels as input, we will get a series of bounding boxes. For each bounding box created by Faster R-CNN, its class is judged.

After the detection of the top view, we get a series of bounding boxes {\it box$_T^1$,box$_T^2$,...,box$_T^m$}. For {\it i}th bounding box {\it box$_T^{\it i}$}, its food type is {\it type$_T^{\it i}$}. Besides these bounding boxes, we regard the bounding box {\it c$_T$} which is judged as the calibration object with the highest score to calculate scale factor of the top view. In the same way, after the detection of the side view, we get a series of bounding boxes {\it box$_S^1$,box$_S^2$,...,box$_S^n$}. For {\it j }th({\it j} $\in${\it 1,2,...,n}) bounding box {\it box$_S^{\it j}$}, its food type is {\it type$_S^{\it j}$}. And we treat the bounding box {\it c$_S$} judged as the calibration object with the highest score to calculate scale factor of the side view.
\subsection{Image Segmentation}
After detection, we need to segment each bounding box. GrabCut is an image segmentation approach based on optimization by graph cuts\cite{grabcut}. Practicing GrabCut needs a bounding box as foreground area which can be provided by Faster R-CNN. Although asking user to label the foreground/background color can get better result, we refuse it so that our system can finish calorie estimation without user’s assistance. For each bounding box, we get precise contour after applying GrabCut algorithm. After segmentation, we get a series of food images {\it P$_T^1$,P$_T^2$,...,P$_T^m$} and {\it P$_S^1$,P$_S^2$,...,P$_S^n$}. The size of {\it P$_T^i$} is the same as the size of {\it box$_T^i$} ({\it i} $\in${\it 1,2,...,m}), but the values of background pixels are replaced with zeros, which means that only the foreground pixels are left. The calibration object boxes {\it c$_T$} and {\it c$_S$} are not applied by GrabCut. After image segmentation, we can estimate every food's volume and calorie.
\subsection{Volume Estimation}
In order to estimate volume of each food, we need to to calculate scale factors based on the calibration objects first. When we use the One Yuan coin  as the reference, according to the coin's real diameter(2.50$\it cm$), we can calculate the side view's scale factor {\it $\alpha_S$}($cm$) with Equation \ref{PS}.
\begin{equation}
\label{PS}
{\it \alpha_S}=\frac{2.5}{(W_S+H_S)/2}
\end{equation}
Where $W_S$ is the width of the bounding box {\it c$_S$} and $H_S$ is the height of the bounding box {\it c$_S$}. 

Then the top view's scale factor {\it $\alpha_T$}($cm$) is calculated with Equation \ref{PT}.
\begin{equation}
\label{PT}
{\it \alpha_T}=\frac{2.5}{(W_T+H_T)/2}
\end{equation}
Where $W_T$ is the width of the bounding box {\it c$_T$} and $H_B$ is the height of the bounding box {\it c$_T$}.  

For each food image {\it P$_T^i$}({\it i} $\in${\it 1,2,...,m}), we try to find an image in set {\it P$_S^1$,P$_S^2$,...,P$_S^n$} with the same food type. If {\it type$_T^{\it i}$} is equal to {\it type$_S^{\it j}$}({\it j} $\in${\it 1,2,...,n}), {\it P$_S^j$} will be marked so that it won't be used again; then {\it P$_T^i$} and {\it P$_S^j$} will be used to calculate this food's volume.
We divide foods into three shape types: ellipsoid, column, irregular. According to the food type {\it type$_T^{\it i}$}, we select the corresponding volume estimation formula as shown in Equation \ref{volume formula}.
\begin{equation}
\label{volume formula}
v = \left\{
             \begin{array}{lcl}
             {\it \beta \times \frac{\pi}{4} \times \sum_{k=1}^{H_S} {(L_S^k)^2 \times \alpha_S^3}} &\text{if} &the \, shape \,  is \, ellipsoid \\
             {\it  \beta  \times (s_T \times \alpha_T^2) \times (H_S \times \alpha_S)} &\text{if} &the \, shape \, is \, column\\  
             {\it  \beta  \times (s_T \times \alpha_T^2) \times \sum_{k=1}^{H_S} {(\frac{L_S^k}{L_S^{MAX}})^2 \times \alpha_S}} &\text{if} &the \, shape \, is\, irregular\\ 
             \end{array}  
        \right.
\end{equation}
In Equation \ref{volume formula}, $H_S$ is the number of rows in side view $\it P_S$ and ${\it L_S^k}$ is the number of foreground pixels in row ${\it k(k \in{ \{ \it 1,2,...,H_S \} }})$. $\it L_S^{MAX}=\max(L_S^1,...,L_S^{H_S})$ records the maximum number of foreground pixels in $\it P_S$. $\it s_T= \sum_{k=1}^{H_T} L_T^k$ is the number of foreground pixels in top view $\it P_T$, where $L_T^k$ is the number of foreground pixels in row ${\it k(k \in{\it 1,2,...,H_T})}$. $\beta$ is the compensation factor and the default value is {\it1.0}. 
\subsection{Calorie Estimation}
After getting volume of a food, we get down to estimate each food's mass first with Equation \ref{mass equation}.
\begin{equation}
\label{mass equation}
{\it m}= \it \rho \times v
\end{equation}
Where $v(cm^3)$ is the volume of current food and $\rho(g/cm^3)$ is its density value.

Finally, each food's calorie is obtained with Equation \ref{calorie equation}.
\begin{equation}
\label{calorie equation}
{\it C}= \it c \times m
\end{equation}
Where $m(g)$ is the mass of current food and $c(Kcal/g)$ is its calories per gram.
\section{Results and Discussion}
\label{experiment}

\subsection{Dataset Description}
For our calorie estimation method, a corresponding image dataset is necessary for evaluation. Several food image datasets\cite{food101,PFID,FooDD,UEC256} have been created so far. But these food datasets can not meet our requirements, so we use our own food dataset named ECUSTFD\footnote{\url{http://pan.baidu.com/s/1o8qDnXC}}.

ECUSTFD contains 19 kinds of food: {\it apple, banana, bread, bun, doughnut, egg, fired dough nut, grape, lemon, litchi, mango, mooncake, orange, peach, pear, plum, qiwi, sachima, tomato}. For a single food portion, we took some pairs of images by using smart phones; each pair of images contains a top view and a side view. For each image, there is only a One Yuan coin as calibration object. If there are two food in the same image, the type of one food is different from another. For every image in ECUSTFD, we provide annotations, volume and mass records.

\subsection{Object Detection Experiment}
\begin{figure}[th]
\includegraphics[width=\textwidth]{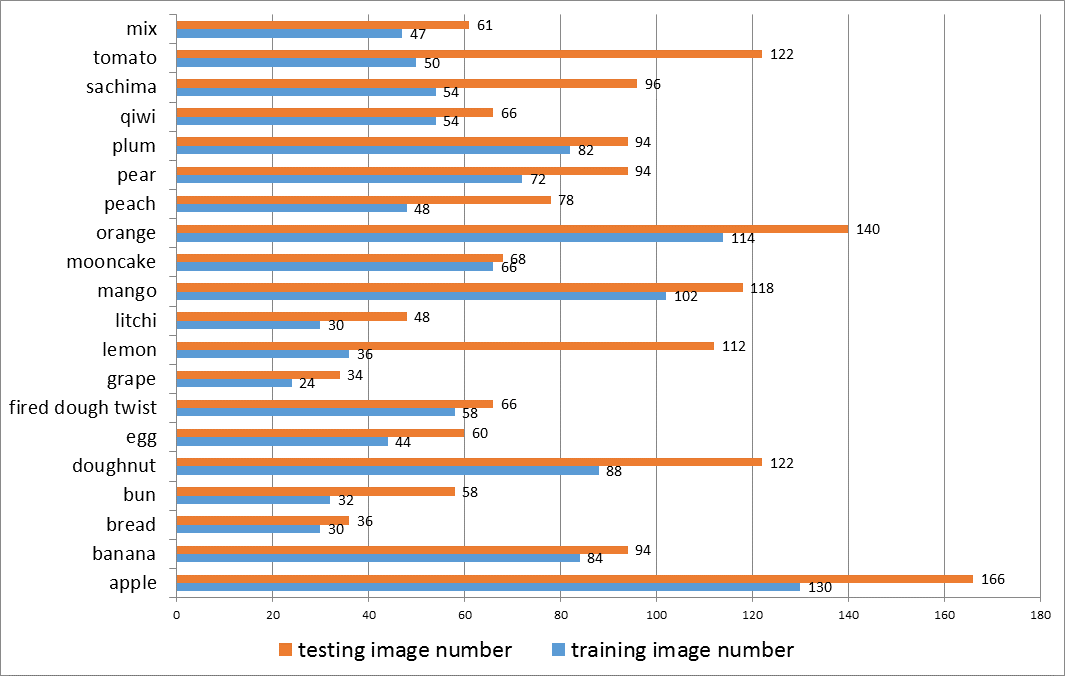}
\caption{training and testing image numbers}
\label{image number}
\end{figure}

In this section, we compare the object detection results between Faster R-CNN and another object detection algorithm named Exemplar SVM(ESVM) in ECUSTFD. In order to avoid using train images to estimate volumes in the following experiments, the images of two sets are not selected randomly but orderly. The numbers of training images and testing images are shown in Figure \ref{image number}. And we use Average Precision(AP) to evaluate the object detection results.
In testing set, Faster R-CNN'S mean Average Precision(mAP) is 93.0\% while ESVM'S mAP is only 75.9\%, which means that Faster R-CNN is up to the standard and can be used to detect object.
\subsection{Food Calorie Estimation Experiment}
In this section, we present our food calorie estimation results. Due to the limit of experimental equipments, we can not get each food's calorie as a reference; so our experiments just verify the volume estimation results and mass estimation results.
First we need to get the compensation factor $\beta$ in Equation \ref{volume formula} and $\rho$ in Equation \ref{mass equation} for each food type with the training sets. $\beta$ is calculated with Equation \ref{beta}.
\begin{equation}
\label{beta}
\beta_k=\frac{\sum_{i=1}^{N}V_k^i}{\sum_{i=1}^{N}v_k^i}
\end{equation}
Where $k$ is the food type, $N$ is the number of volume estimation. $V_k^i$ is the real volume of food in the $i$th volume estimation; and $v_k^i$ is the estimation volume of food in the $i$th volume estimation.

$\rho$ is calculated with Equation \ref{rho}.
\begin{equation}
\label{rho}
\rho_k=(\frac{\sum_{i=1}^{N}M_k^i}{\sum_{i=1}^{N}V_k^i})
\end{equation}
Where $k$ is the food type, $N$ is the number of mass estimation. $M_k^i$ is the real mass of food and $V_k^i$ is the real volume of food in the $i$th mass estimation.

The shape definition, estimation images number, $\beta$, $\rho$ of each food type are shown in Table \ref{Parameters}. For example, we use 122 images to calculate parameters for apple, which means that $N=122/2=61$ volume estimation results are used to calculate $\beta$.
\begin{table}[htb!]
\caption{Shape Definition and Parameters in Our Experiments}
\label{Parameters}
\centering
{\begin{tabular}{|c|c|c|c|c|c|} 
\hline
Food Type & shape &  estimation image& $\beta_k$ & $\rho_k$ \\
&&  number&&\\
\hline
apple &ellipsoid& 122 & 1.08 &0.78  \\ 
banana &irregular&  82 & 0.62 &0.91  \\ 
bread &column& 26 & 0.62 &0.18 \\ 
bun &irregular& 32 & 1.07 &0.38 \\ 
doughnut &irregular&  42 & 1.28 &0.30  \\ 
egg &ellipsoid&  30 & 1.01 &1.17 \\ 
fired dough twist&irregular& 48 & 1.22 &0.60 \\ 
grape &column&  24 & 0.45 &1.00 \\ 
lemon &ellipsoid&  34 & 1.06 &0.94 \\ 
litchi  &irregular&  30 & 0.82 &0.98 \\ 
mango &irregular&  20 & 1.16 &1.08 \\ 
mooncake &column&  64 & 1.00 &1.20 \\ 
orange &ellipsoid&  110 & 1.09 &0.88 \\ 
peach &ellipsoid& 48 & 1.05 &1.01 \\ 
pear &irregular&  72 & 1.02 &0.97 \\ 
plum &ellipsoid&  82 & 1.22 &0.97\\ 
qiwi &ellipsoid&  54 & 1.16 &0.98  \\ 
sachima &column&  54 & 1.10 &0.22 \\ 
tomato &ellipsoid&  46 & 1.21 &0.90   \\ 
\hline
\end{tabular}}
\end{table}

\begin{table}[htb!]
\caption{Volume and Mass Estimation Experiment Results}
\label{testing Results}
\centering
\addtolength{\tabcolsep}{-6pt}
{\begin{tabular}{|c|c|c|c|c|c|c|c|c|} 
\hline
Food Type&estimation&mean& mean&volume&mean&mean &mass \\
&  image & volume& estimation& error& mass&estimation& error\\
&  number && volume&(\%)&&mass&(\%)\\
\hline
apple & 154 & 333.64 & 270.66 & -11.59& 263.82 & 292.51 & -13.14 \\ 
banana & 90 & 162.00 & 204.16 & -21.42& 146.38 & 127.24 & -20.94 \\ 
bread & 20 & 155.00 & 180.62 & -26.47& 29.04 & 112.04 & -29.13 \\ 
bun & 56 & 245.36 & 235.39 & 2.80& 77.87 & 252.11 & 22.76 \\ 
doughnut & 118 & 174.75 & 143.47 & 5.36& 56.26 & 183.64 & -0.74 \\ 
egg & 34 & 52.94 & 62.13 & 17.56& 61.64 & 62.52 & 17.67 \\ 
fired dough twist & 44 & 65.00 & 54.50 & 4.78& 40.80 & 66.64 & -2.53 \\ 
grape & 30 & 240.00 & 323.57 & -38.86& 219.50 & 146.73 & -33.45 \\ 
lemon & 112 & 96.79 & 94.03 & 3.88& 94.24 & 100.11 & 0.13 \\ 
litchi & 48 & 43.33 & 49.25 & -6.05& 43.80 & 40.54 & -8.88 \\ 
mango & 96 & 81.67 & 70.43 & 4.34& 88.11 & 81.48 & -0.07 \\ 
mooncake & 66 & 67.58 & 52.52 & -16.57& 62.13 & 52.54 & 2.15 \\ 
orange & 104 & 234.42 & 235.93 & 10.06& 216.99 & 257.83 & 4.66 \\ 
peach & 62 & 110.65 & 115.52 & 6.46& 117.06 & 121.03 & 1.62 \\ 
pear & 82 & 260.00 & 225.93 & -10.05& 248.10 & 229.41 & -8.91 \\ 
plum & 94 & 100.00 & 98.35 & 20.22& 105.14 & 120.22 & 11.35 \\ 
qiwi & 56 & 127.14 & 123.08 & 8.65& 127.31 & 143.33 & 5.93 \\ 
sachima & 96 & 147.29 & 129.05 & -3.28& 31.89 & 142.35 & -1.48 \\ 
tomato & 90 & 174.22 & 168.28 & 17.11& 182.64 & 203.07 & 0.36 \\ 
\hline
\end{tabular}}
\end{table}
Then we use the images from the testing set to evaluate the volume and mass estimation results. The results are shown in Table \ref{testing Results} either. We use mean volume error to evaluate volume estimation results. Mean volume error is defined as:
\begin{equation}
\label{ME}
{\it ME_V^i}=\frac{1}{N_i} \sum_{j=1}^{N_i} \frac{v_j-\uppercase{V}_j}{\uppercase{V}_j}
\end{equation}
In Equation \ref{ME}, for food type ${i}$ , $2{N_i}$ is the number of images Faster R-CNN recognizes correctly. Since we use two images to calculate volume, so the number of estimation volumes  for type ${i}$ is ${N_i}$. ${v_j}$ is the estimation volume for the ${j}$th pair of images with the food type ${i}$; and ${\uppercase{V}_j}$ is corresponding real volume for the same pair of images.
Mean mass error is defined as:
\begin{equation}
\label{MME}
{\it ME_M^i}=\frac{1}{N_i} \sum_{j=1}^{N_i} \frac{m_j-\uppercase{M}_j}{\uppercase{M}_j}
\end{equation}
In Equation \ref{MME}, for food type ${i}$ , the number of mass estimation  for ${i}$th  type is ${N_i}$. ${m_j}$ is the estimation volume for the ${j}$th pair of images with the food type ${i}$; and ${\uppercase{M}_j}$ is corresponding real mass for the same pair of images.

Volume and mass estimation results are shown in Table\ref{testing Results}. For most types of food in our experiment, the estimation results are closer to reference real values. The mean error between estimation volume and true volume does not exceed ${\pm}$20\% except banana, bread, grape, plum. For some food types such as lemon, our estimation result is close enough to the true value. The mass estimation results are almost the same as the volume estimation results. But for some food types like mooncake and tomato, the mass estimation errors are less than the volume estimation errors; the way we measure volume needs to be blamed due to drainage method is not accurate enough. All in all, our estimation method is available.
\section{CONCLUSION}
\label{conclusion}
In this paper, we provided our calorie estimation method. Our method needs a top view and side view as its inputs. Faster R-CNN is used to detect the food and calibration object. GrabCut algorithm is used to get each food's contour. Then the volume is estimated with volume estimation formulas. Finally we estimate each food's calorie. The experiment results show our method is effective. 
\section*{REFERENCE}
\bibliographystyle{unsrt}
\bibliography{sample}

\end{document}